\documentclass[12pt]{article}
\usepackage{fullpage}
\usepackage{subcaption}
\usepackage{graphicx}
\usepackage{natbib}
\usepackage{amsmath}
\usepackage{amsfonts}
\usepackage{authblk}

\newcommand{\rulesep}{\unskip\ \vrule\ }

\begin{document}

\title{Evaluating Reinforcement Learning Algorithms \\ in Observational Health Settings}
\author[1]{Omer Gottesman}
\author[2]{Fredrik Johansson}
\author[1]{Joshua Meier}
\author[1]{Jack Dent}
\author[1]{Donghun Lee}
\author[1]{Srivatsan Srinivasan}
\author[3]{Linying Zhang}
\author[3]{Yi Ding}
\author[1]{David Wihl}
\author[1]{Xuefeng Peng}
\author[1]{Jiayu Yao}
\author[1]{Isaac Lage}
\author[4]{Christopher Mosch}
\author[2]{Li-wei H. Lehman}
\author[5,6]{Matthieu Komorowski}
\author[7]{Aldo Faisal}
\author[5,8,9]{Leo Anthony Celi}
\author[2]{David Sontag}
\author[1]{Finale Doshi-Velez}

\affil[1]{Paulson School of Engineering and Applied Sciences, Harvard University}
\affil[2]{Institute for Medical Engineering and Science, MIT}
\affil[3]{T.H. Chan School of Public Health, Harvard University}
\affil[4]{Department of Statistics, Harvard University}
\affil[5]{Laboratory for Computational Physiology, Harvard-MIT Health Sciences \& Technology, MIT}
\affil[6]{Department of Surgery and Cancer, Faculty of Medicine, Imperial College London}
\affil[7]{Department of Bioengineering, Imperial College London}
\affil[8]{Division of Pulmonary, Critical Care and Sleep Medicine, Beth Israel Deaconess Medical Center}
\affil[9]{MIT Critical Data}

\date{}
\maketitle

\section{Motivation}
Much attention has been devoted recently to the development of machine learning algorithms with the goal of improving treatment policies in healthcare.  Reinforcement learning (RL) is a sub-field within machine learning that is concerned with learning how to make sequences of decisions so as to optimize long-term effects. Already, RL algorithms have been proposed to identify decision-making strategies for mechanical ventilation~\citep{prasad2017reinforcement}, sepsis management~\citep{raghu2017deep} and treatment of schizophrenia~\citep{shortreed2011informing}.  However, before implementing treatment policies learned by black-box algorithms in high-stakes clinical decision problems, special care must be taken in the evaluation of these policies. 

In this document, our goal is to expose some of the subtleties associated with evaluating RL algorithms in healthcare.  Specifically, we focus on the \emph{observational} setting, that is, the setting in which our RL algorithm has proposed some treatment policy, and we want to evaluate it based on historical data.  This setting is common in healthcare applications, where we do not wish to experiment with patients' lives without evidence that the proposed treatment strategy may be better than current practice.  

While formal statistical methods have been developed to assess the quality of new policies based on observational data alone \citep{thomas2016data,precup2000eligibility,pearl2009causality,imbens2015causal}, these methods rely on strong assumptions and are limited by statistical properties.  We do not attempt to summarize this vast literature in this work, rather, we aim to provide a conceptual starting point for clinical and computational researchers to ask the right questions when designing and evaluating algorithms for new ways of treating patients.  In the following, we describe how choices about how to summarize a history, variance of statistical estimators, and confounders in more ad-hoc measures can result in unreliable, even misleading estimates of the quality of a treatment policy.  We also provide suggestions for mitigating these effects---for while there is much promise for mining observational health data to uncover better treatment policies, evaluation must be performed thoughtfully. 

% One such assumption is unconfoundedness (ignorability)---that all variables affecting both observed treatments and outcomes have been measured. This assumption may be broken any time a care provider makes a decision based on information that is predictive of an outcome, but not recorded in data. As a result, depending on the application, the unconfoundedness assumption may be hard to justify when learning only from electronic health record (EHR) data. Moreover, if the policy under evaluation differs significantly from the observed policy, the statistical power of standard evaluation methods is significantly reduced. For these reasons, 

\section{Running Example: Managing Sepsis in the ICU}
\label{sec:setting}

There exist many healthcare scenarios in which treatment involves a series of decisions. To make the remainder of the document concrete, we will use sepsis management as our running example.  Below, we first describe the sepsis management setting and then describe how we formalize the clinical problem into the reinforcement learning framework.

Specifically, we consider decisions around when and how much vasopressor and IV-fluids should be administered to patients with sepsis.  In this setting, identifying a good treatment strategy is challenging because the true outcome of interest, mortality, may not be observed for days after a series of decisions are made, and shorter term goals (such as maintaining a particular blood pressure level), may not necessarily correlate strongly with better mortality outcomes.  Patients in critical care settings also generate a large amount of data, and it may not be immediately obvious what parts of a patient's history are most relevant for making the decision.  

We use a cohort of sepsis patients curated from the MIMIC III dataset~\citep{johnson2017mimic,raghu2017deep}. The dataset consists of 47 features (including demographics, vitals, and lab test results) measured on 19,275 ICU patients who fulfilled the sepsis-3 definition criteria, namely the  presence of a suspected infection along with evidence of organ dysfunction \citep{singer2016third}. Following \citet{raghu2017deep}, for every patient, each parameter is aggregated at 4-hour intervals. We also record administration of IV-fluids and vasopressors, as well as mortality within 90 days of admission to the ICU.  Our goal is to determine, at every 4-hour interval, whether and how much vasopressor and IV fluid should be administered.  This problem is challenging because these treatments might have long term effects such as interstitial fluid accumulation and subsequent organ dysfunction following excessive fluid administration.

To apply reinforcement learning techniques to the sepis management problem, we need to define three things: a state space, an action space, and a reward function.  The \emph{state} representation $s$ can be thought of all the relevant covariates about a patient at some point in time; these are the inputs to the decision-making system.  In general,  for each patient, we have a \emph{history} $H$ of their measurements and treatments to date.  We define a \emph{policy} $a=\pi{H}$ as a mapping from the patient's history to the next treatment decision.  However, working with entire histories can be unweildy; and so reinforcement learning practioners often simplify the problem by defining a simpler set of covariates $s$ to summarize the history.  In Section~\ref{sec:representation}, we discuss how the choice of summary statistic can impact the quality of the policy learned.

% Patients with similar characteristics and severity are grouped into homogeneous health states. During the course of their stay in the ICU, a patient can visit several states as their disease progresses. We cluster the data into $K$ clusters using unsupervised clustering and assume that the treatment decision can be made just based upon the patient's current cluster.  In reinforcement learning, this simplification is known as the Markovian assumption, and the cluster as the patient's \emph{state}, $s$.

For the action (or treatment) space, we also make some simplifying assumptions.  We discretize each of our treatments (IV fluids and vasopressors) into 5 bins, the first representing no treatment (zero dosage), and the rest representing quartiles of the actions prescribed by physicians. The combination of these two actions results in 25 possible treatment choices that we must choose from for every 4-hour interval.  Focusing on IV fluids and vasopressors assumes that those are the only treatments that we have control over; the patient may receive other treatments (e.g. antibiotics) but those are out of our scope. 

Finally, we must define a reward function.  In reinforcement learning, the reward $r$ represents the desirability of the immediate outcomes of the action taken, and the goal of an RL policy is to learn the policy which will maximize for all patients the total reward collected over the entire history, $R=\sum_{t=0}^T r_t$, where $r_t$ is the reward received at time $t$, and $T$ is the total number of time intervals in the patient's history. In our analyses, we assume that our decision-making agent receives a reward of zero up until its last action; at the final time interval of the history the agent receives a positive reward if the patient survives and a negative reward otherwise.

Once the state, action, and reward structure of the problem have been formalized, we are ready to start searching for an optimal policy $\pi$.  In reinforcement learning, the quality of a policy is quantified by the expected cumulative rewards, known as the \emph{value} $V^\pi = \mathbb{E}_{H\sim \pi}[R^H]$.  If we could implement our treatment policy, calculating the above would be easy: we would simply try our policy on many patients and then compute the average returns (outcomes).  However, in many cases, we cannot directly experiment with patients in this way; we first want to estimate how well a policy might perform given historical data.  The challenge of computing the value of one policy given another is known as \emph{off-policy evaluation} and is the subject of this discussion.

% FDV: Maybe incorporate into the above if we think it's important, I don't think so... 
% learn a table-based MDP and solve it to obtain an optimal policy~\citep{sutton1998reinforcement}.

\section{Why do we need careful evaluation?}

Before diving into the sublteties associated with off-policy evaluation, we first provide two illustrative examples of why careful evaluation is needed in the first place.  

For the first illustration, we describe what happens when we apply a standard reinforcement learning algorithm to the sepsis management problem above.  We observed a tendency of learned policies to recommend minimal treatment for patients with very high acuity (SOFA score). This recommendation makes little sense from a clinical perspective, but can be understood from the perspetive of the RL algorithm. Most patients who have high SOFA score receive aggressive treatment, and because of the severity of their condition, the mortality rate for this subpopulation is also high.  In the absence of data on patients with high acuity that received no treatment, the RL  algorithm concludes that trying something rarely or never performed may be better than treatments known to have poor outcomes.

For our second illustration, we consider what happens when the RL agent is tasked to \emph{only} consider when to intubate a patient (and not provided agency over other treatments, such as IV fluids or vasopressors).  Limiting the number of treatments makes the RL problem more tractable, but we observed that the RL agent recommended intubation noticably more than clinicians.  Upon further inspection, it turns out that the overuse of intubation is a result of a lack of treatment options: the RL algorithm learns to recognize patients who need additional care, and, in the absence of other options, takes the only action available to it.  When we trained an RL agent with more treatment choices (mechanical ventillation, IV fluids, and vasopressors), the overuse of mechanical ventillation disappears.

The susceptibility of AI algorithms to learn harmful policies due to artifacts in the data or methodology highlights the importance of evaluation. In the following sections we demonstrate how and when standard evaluation methods may fail to recognize such artifacts, and recommend best practices for evaluation.

\section{Challenges with Choosing Representations}
\label{sec:representation}

\paragraph{The Challenge}
As we mentioned in our overview of the sepsis management problem, the first step in applying the reinforcement learning framework to a problem is determining how to summarize a patient's history $H$ into some more manageable covariate that we define as the state representation $s$.  In machine learning, this task is known is \emph{choosing a representation}, because it involves deciding how to represent a patient's history in a way that retains sufficient information for the task at hand.  It is essential that this representation accounts for any variables that might \emph{confound} estimates of outcomes under the policy~\citep{robins2000robust}, that is, we must include any factors that causally affect both observed treatment decisions and the outcome.  For example, sicker patients receive more aggressive treatments and have higher mortality.  If one fails to adjust for the acuity of the patient, one might erroneously conclude that treatment increases risk of mortality, even when the opposite is true. 

Formally, confounding variables represent any information that influenced a decision \emph{and} had an effect on the outcome of that decision. These variables may include a patient’s entire medical record, their genetic make-up, as well as socio-economic and environmental factors. Unfortunately, it is impossible to verify that all confounders have been measured based on statistical quantities alone~\citep{robins2000robust}. Instead, domain knowledge must be applied to justify this claim.  Does the clinical researcher, based on their knowledge of the disease, believe that all important factors have been measured?  We note that identifying confounding factors is a substantial task even for the simpler problem of making a single treatment decision, and this problem is only aggravated in the sequential setting~\citep{chakraborty2014dynamic}.

Moreover, even if all potential confounders have been measured, it is important to ensure that the data structure passed to the RL algorithm retains this information. In particular, as we mentioned before, a patient's history is a large and often unwieldy object, and thus it is standard practice to compress this history into some features known as the state representation. All RL algorithms assume that the features given to them are sufficient summaries of the entire patient history and will fail if the representation is insufficient. However, identifying such a representation is not straightforward. For example, to make a prediction about a patient's response to a specific medication such as a vasopressor in the ICU, is it sufficient to recall the last dose administered? The cumulative dose given throughout the ICU stay? Each instance it was given?  The correct choice of summary will depend on the specific clinical problem at hand.  

% Below we describe a simple example demonstrating how sensitive RL algorithms could be to the choice of representation. We implement the simple approach described in Section~\ref{sec:setting}, of clustering patients into discrete states using $k$-means clustering. Such a choice of data discretization can dramatically affect results and conclusions, as we show below. Apart from diminishing information content of the data by effectively performing a dimensionality reduction, clustering algorithms are highly sensitive to initial conditions, which may be randomized (such as in the case of $k$-means) or difficult to choose.

\paragraph{Illustration}
To illustrate the sensitivity of RL algorithms to these kinds of representation choices, we decided to use a simple clustering algorithm, k-means, to divide the patient's current measurements into $K=100$ clusters.  At any point in time, we assumed that the patient's current cluster was sufficient to summarize their history, and then used the cluster as our covariate to learn a sepsis treatment policy---clearly a simplified view of sepsis, but potentially the innocent kind of simplification we can all imagine making. Next, we repeated the process four times with different random initializations to the clustering algorithm.  Across these initializations, the agreement between treatment choices from the four runs and the initial run was only $26\pm5\%$.  If we performed the same experiment with $K=200$ clusters and compared the treatment recommendations learned to our base recommendation ($K=100$), the agreement on treatment with the base recommendation (with $K=100$) drops to $14\pm2\%$.  This low agreement, even when running the same algorithms with the same data, highlights the need to choose a meaningful representation.

\paragraph{Diagnosing and Mitigating Concerns}
While there are some ways in which to ensure that the representation that is learned is adequate (e.g. \cite{shalit2016estimating}), it is ultimately not possible to ascertain, via quantitive means, whether the data contain all the confounders.  When quantitive evaluation is challenging, we must rely on other methods to assess the validity of our ideas. Interpretability is one fall-back when we don't have quantitative options: one should always ask expert clinicians whether the final state representation seems reasonable for decision-making or is missing important information.

\section{Challenges with Statistical Evaluation}
\label{sec:IS_estimators}

\paragraph{Challenges}
Once we have a state representation (Section~\ref{sec:representation}, there is still the question of how to evaluate the treatment policy recommended by the RL algorithm given only observational data.  A large class of off-policy evaluation methods are built on the principle of \emph{importance sampling} (IS). At their core, given a large batch of retrospective data observed under the clinician policy $\pi_b$, these methods assign weights to individual samples to make the data appear to be drawn from the evaluation policy, $\pi_e$. The basic importance sampling estimator of $V^{\pi_e}$ is given by:
\begin{equation}
\hat{V}^{\pi_{e}} = \frac{1}{N}\sum_{n=1}^{N}w^{H_n}R^{H_n},
\label{eq:is}
\end{equation}
where $H_n$ denotes the history of patient $n$, and $N$ is the total number of patients.

The weights $w^{H_n}$ in \eqref{eq:is} tell us how important the $n^{th}$ sample is for estimating $V^{\pi_e}$. In the basic IS estimator, this corresponds to how likely it is to observe the history $H_n$ is under the RL-based policy compared to how likely it is under the clinician policy:
\begin{equation}
w^{H_n} = \prod_{t=0}^{T_{H_n}}\frac{\pi_{e}(a_t^{H_n}|s_t^{H_n})}{\pi_{b}(a_t^{H_n}|s_t^{H_n})},
\end{equation}
with $T_{H_n}$ being the number of time steps in history $n$, and $\pi_{e}(a_t^{H_n}|s_t^{H_n})$ and $\pi_{b}(a_t^{H_n}|s_t^{H_n})$ being the probabilities of taking at time $t$ action $a_t^{H_n}$ given state representation $s_t^{H_n}$ under the RL algorithm's and clinician's policies, respectively. Intuitively, when the clinician takes actions that the RL algorithm does not, the weight attributed to that trajectory and outcome will be small.  In this way, IS-based methods can be thought of as sub-sampling the trajectories from the observational data that match the treatment policy recommended by the RL algorithm.

A short calculation shows that the IS estimator is an unbiased estimator of $V^{\pi_e}$---given enough samples, we obtain the correct value~\citep{precup2000eligibility}. However, when the evaluation policy is deterministic, i.e. $\pi_e(a\mid s)$ is one for one action and zero for all others, the only histories with non-zero weight are those in which all observed actions match the actions proposed by the evaluation policy---all other samples are effectively discarded. As a result, the number of informative samples may be very small if $\pi_e$ and $\pi_b$ differ significantly, and the variance of the IS estimator will be high.
To overcome this, there exists a large body of work on striking beneficial tradeoffs between bias and variance in off-policy evaluation estimators \citep{precup2000eligibility, jiang2015doubly, thomas2016data}. The details of each estimator are given in the provided references, but the point we emphasize here is that \emph{all} of them are unreliable for evaluation in the sepsis management problem for two main reasons:
\begin{itemize}
	\item Outcomes are sparse in time: our histories can have up to 20 time-steps in which decisions are made, but we only see the outcome (mortality) at the end.  (In contrast, a problem involving glucose management in diabetes might involve a reward for staying in range at every time step.)  Having sparse outcomes makes it challenging to assess the value of an individual treatment decision; often, we can only make assessments about a collection of decisions.  
	\item Unless the (deterministic) policy we wish to evaluate is very similar to the one already used by physicians, most of the weights will be zero---only a tiny fraction of histories in the data will have all the treatments that our policy recommends.  In the absence of an intermediate reward signal, we need to find entire sequences that match to uncover whether the treatment was successful.
\end{itemize}
The combination of these properties results in few samples being assigned non-zero weight, which in turn results in high-variance estimates of the quality of the evaluation policy.

\paragraph{Illustration}
We now illustrate the consequences of the challenges above in our sepsis management case study.  We test the performance of four modern importance sampling methods on the dataset described in Section~\ref{sec:setting}: per-decision importance sampling (PDIS), weighted per-decision importance sampling (WPDIS) \citep{precup2000eligibility}, doubly-robust (DR)~\citep{jiang2015doubly}, and weighted doubly-robust (WDR)~\citep{thomas2016data}. The weighted methods trade increased bias for reduced variance, while the per-decision methods reduce variance by computing the weights $w^{H_n}$ in a way that does not penalize the probability of a current action based on future ones. Doubly robust methods leverage an approximate model of the reward function to reduce variance.

In the experiments below, we defined the state representation based on clustering the patient's  current observations into $K=750$ clusters.  This definition was kept constant throughout the experiments.  Next, we randomly partitioned the data 50 times into training sets (80\% of the data) and test sets (20\%) and learned a treatment policy based on the training set.  We then estimated the value of this optimal policy with the test data using the off-policy estimators described earlier. For every partitioning, we also computed a \emph{model-based} value estimate by building a model from the training data.  Model-based estimates are biased, but often have less variance than IS-based estimators. 

\begin{figure}[ht]
	\centering
	\begin{subfigure}{0.4\linewidth}
		\centering
		\includegraphics[width=\linewidth]{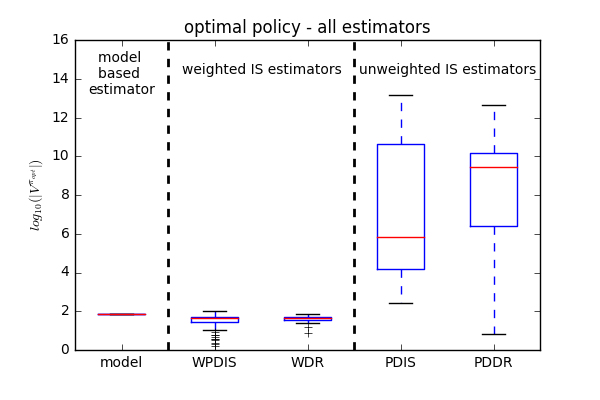}
		\caption{}
		\label{subfig:IS_estimators_boxplot_full}
	\end{subfigure}
	\rulesep
	\begin{subfigure}{0.4\linewidth}
		\centering
		\includegraphics[width=\linewidth]{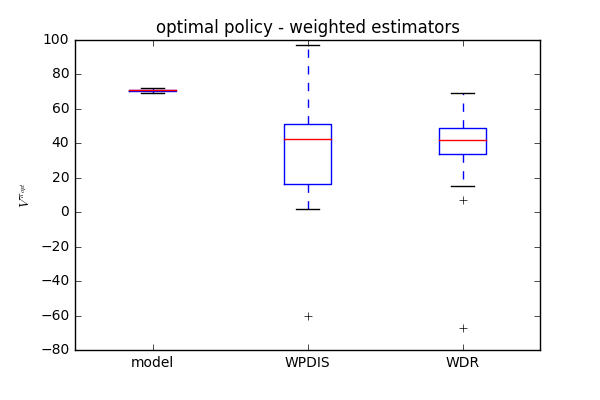}
		\caption{}
		\label{subfig:IS_estimators_boxplot}
	\end{subfigure}
	\begin{subfigure}{0.4\linewidth}
		\centering
		\includegraphics[width=\linewidth]{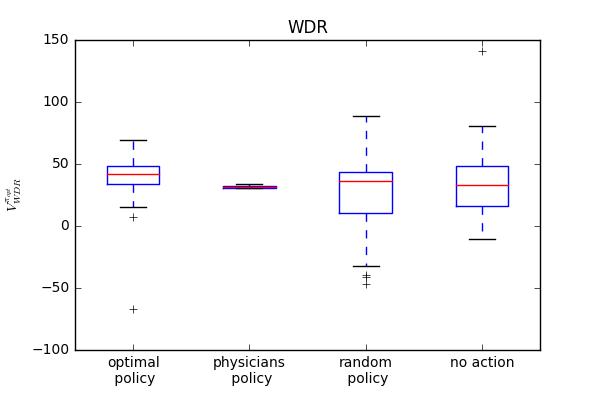}
		\caption{}
		\label{subfig:WDR_estimators_boxplot}
	\end{subfigure}
	\rulesep
	\begin{subfigure}{0.4\linewidth}
		\centering
		\includegraphics[width=\linewidth]{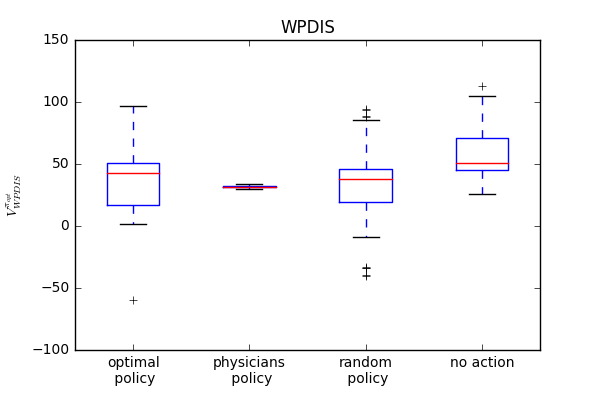}
		\caption{}
		\label{subfig:WPDIS_estimators_boxplot}
	\end{subfigure}
	\caption{Policy value estimates. All boxes represent results over 50 random partitionings of the data into training (80\%) and test (20\%) set. (a) Estimates (log of the absolute value) of the value of the optimal policy learned by our model using different estimators. The unweighted estimators give results which are unrealistic by many orders of magnitude. (b) True values (not logged) of a subset of the data shown in (a). The IS estimators have a large variance, but give more realistic estimates than the model-based estimator. (c-d) Estimates of the values of different policies using the WDR (c) and WPDIS (d) estimators. While the results are noisy and inconclusive, the WDR estimator suggests that the model-based optimal policy outperforms other policies.}
	\label{fig:IS_estimators_boxplots}
\end{figure}

Figure~\ref{fig:IS_estimators_boxplots} shows box plots representing the distribution of values of the estimators obtained for different random partitionings of the cohort (top) and for different policies (bottom). We observe (Figure~\ref{subfig:IS_estimators_boxplot_full}) that the unweighted estimators have very large variance---\emph{many orders of magnitude larger than the $\pm100$ possible range in which the actual value of the policies might lie}. This is typical of the unweighted versions of the IS algorithms when data is too sparse, and we therefore turn our attention to the weighted estimators, in spite of the bias they introduce \citep{precup2000eligibility}. It is important to point out that such large variance in unweighted estimators raises red flags about placing too much confidence in the weighted version of the estimators as well.  The variance of the model-based estimate, as expected, is much smaller (Figure~\ref{subfig:IS_estimators_boxplot}), but is quite optimistically biased\footnote{For a discount factor of $\gamma=0.95$, average sequence length of $\bar{T}=13$ time-steps, $+100$ reward for a patient's survival and $-100$ for death, an estimate of the value of a policy $V^{\pi}$ as a function of cohort mortality rate, $m$, is $V^{\pi}\approx\gamma^{\bar{T}}(-100\cdot m + 100\cdot(1-m))$. This value may be made more accurate by taking into account the full distribution of patients ICU stay lengths for both survivors and non-survivors. For our cohort mortality of $21\%$ the expected value of the physicians' policy is 30, which is consistent with the results shown in Figure~\ref{subfig:WDR_estimators_boxplot}. The model-based value approximation of 50 already corresponds to a zero mortality rate, and the model-based approximation of $V^{\pi}=70$ can be thought of the model expecting to have zero mortality rate and being able to reduce ICU stay to less than 6 time-steps on average.}. The weighted importance sampling methods give more realistic estimates of the policies values, and, as expected, the WDR method has smaller variance than the WPDIS, even though both of them have approximately the same mean.

To understand if these importance sampling methods are useful for policy evaluation we investigate whether they allow us to confidently differentiate between the performance of different policies. In Figure~\ref{subfig:WDR_estimators_boxplot} we compare the WDR estimates for the values of four different policies: the RL-based policy from our model, the physicians' behavior policy, a policy which takes actions uniformly at random, and a policy of never giving any treatment. The value of the physician's policy is estimated with high confidence, which is expected, as in this case our evaluation reduces to an \emph{on-policy} evaluation. The RL-based policy is the only one for which the bounds on the estimated value is relatively tight and larger than the value of the physicians' policy. ($1^{st}$ to $3^{rd}$ quartile box small and above the mean for the physician's policy, in contrast to the large boxes extending to well below the physicians' policies for the random and no-action policies).  That said, all of the policies have relatively close median values and large variances, making it hard to draw definitive conclusions.  Moreover, because the WDR estimator uses a model to reduce variance, it also inherits the optimistic bias of the model---note that the model-free WPDIS estimator in Figure~\ref{subfig:WPDIS_estimators_boxplot} has larger variances and the possible dominance of the RL algorithm's policy has vanished.

%To try and avoid this model bias, it is possible to learn two models, one from the training data to learn an optimal policy, and one form the test data which is used for evaluation. It should be noted however, that unless a large set of the data is used for testing, the models learned for evaluation can be very noisy. (I ACTUALLY TRIED THAT AND INDEED THE MODELS I LEARN WITH ONLY 20\% OF THE DATA ARE SUPER NOISY SO THIS IDEA JUST MAKES EVALUATION WORSE).

The estimates of the model-free WPDIS estimator in Figure~\ref{subfig:WPDIS_estimators_boxplot} also demonstrate how unexpected biases can creep into the IS methods. The surprising feature in Figure~\ref{subfig:WPDIS_estimators_boxplot} is that the policy of never treating patients seems to be outperforming all other treatment policies. This phenonoma occurs because the numerator of the importance weights $w^{H_n}$ is a product of $\pi_e(a_t^{H_n}|s_t^{H_n})$, which, for a deterministic policy, is a product of 0's and 1's. The only sequences retained in the estimate are ones in which the patient receives no (IV fluid or vasopressor) treatment.  These patients form a lower risk group, and therefore the WPDIS method suffers from a selection bias which causes it to predict a much lower mortality rate under the no-treatment policy.

\paragraph{Diagnosing and Mitigating Concerns} 

\begin{figure}[ht]
  \centering
  \begin{subfigure}{0.45\linewidth}
    \centering
    \includegraphics[width=\linewidth]{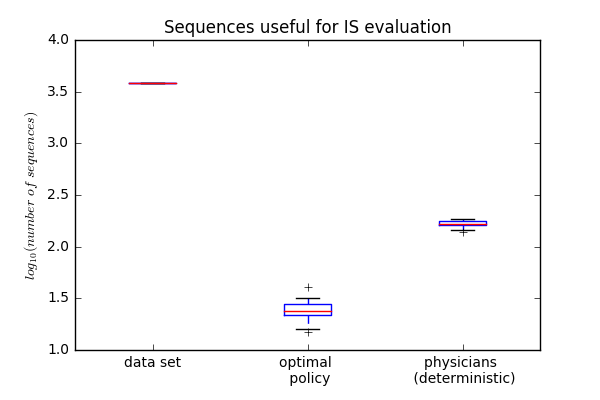}
    \caption{}
    \label{subfig:num_of_usefull_seqs}
  \end{subfigure}
  \rulesep
  \begin{subfigure}{0.45\linewidth}
    \centering
    \includegraphics[width=\linewidth]{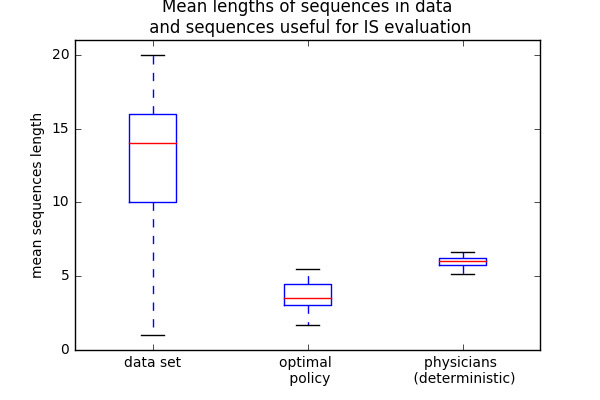}
    \caption{}
    \label{subfig:length_of_usefull_seqs}
  \end{subfigure}
  \caption{Analysis of useful sequences for IS evaluation. (a) The log of the number of sequences in the data compared with the number of sequences for which the physicians action are in full agreement with the actions recommended by the optimal policy from the model, and from a deterministic policy which always takes the most common action used by physicians. (b) Mean lengths of the sequences found in (a). As long as we try to evaluate a deterministic policy, even in the best case scenario the IS methods use only a small fraction of patients' data, and these are usually uncharacteristic patients with very short ICU stays.}
  \label{fig:analysis_of_useful_sequences}
\end{figure}

The illustration above emphasizes one of the most important practice recommendations when using any of the importance sampling methods: \emph{always examine the distribution of the importance weights.} We demonstrate such analysis in Figure~\ref{fig:analysis_of_useful_sequences}. For our data, for example, the test data includes 3855 patients. However, the average effective sample size, i.e. the number of patients for which the importance weight is non-zero under the learned policy is only 25. Furthermore, the average length of stay of such patients is 3.7 time-intervals, whereas the average ICU stay in the data is 13 time-intervals. Even for a policy which takes in every state the action most commonly used by physicians (i.e. the deterministic policy which is as close as possible to the physicians'), the average effective sample size is only 167 out of the 3855 patients in the cohort, and the average ICU stay for these patients is 6 time-intervals. These results demonstrate that even in the best case scenario only a very small fraction of the data is actually used in the IS methods, and examining the distribution of importance weights can give us a sense of how much the IS estimates can be trusted.

A way to increase the robustness of IS estimators is to evaluate non-deterministic policies. If more than one action per state has a non-zero probability of being recommended, the fraction of the data which is attached to non-zero importance weights will be significantly increased. In terms of application of these policies to health-care, given a patient's state, a physician will see a variety of possible treatment recommendation, each with its own level of confidence, and the clinician will need to use their own judgment to choose between the most recommended actions. Having clinicians exercise their own judgment in whether or not to adopt an algorithm suggestion might be desirable in its own right, and therefore introducing such a constraint on the learned policy may be beneficial not only for the purpose of evaluation.

\section{Challenges With Ad-Hoc Evaluation Measures}

\paragraph{Challenge}
Due to both the mathematical sophistication and data-related concerns associated with statistical off-policy evaluation, other works choose to use more direct measures of policy performance. In this section, we consider a class of measures that we call the U-curve method~\citep{raghu2017deep,prasad2017reinforcement}, which is based on the idea of associating the difference between the clinician's policy and the evaluation policy with some outcome, such as mortality.  If this association is positive, the reasoning goes, then it must mean that when there is no difference---that is, when the clinician's action agreed with the suggested action---outcomes were the best.

\begin{figure}[ht]
	\centering
	\begin{subfigure}{0.45\linewidth}
		\centering
		\includegraphics[width=\linewidth]{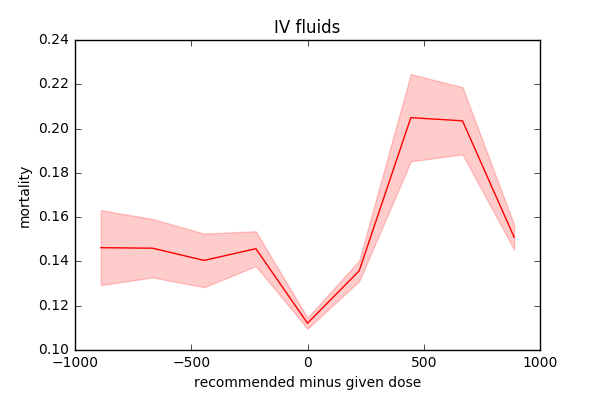}
	\end{subfigure}
	\rulesep
	\begin{subfigure}{0.45\linewidth}
		\centering
		\includegraphics[width=\linewidth]{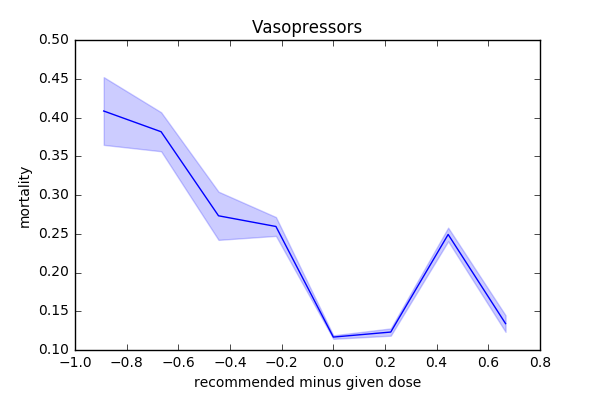}
	\end{subfigure}
	\caption{Mortality vs. deviation from the dose recommended by the optimal policy learned from the model. Without careful analysis, the characteristic U-shaped plots seem to predict the learned policy to significantly outperform the clinicians' policy.}
	\label{fig:u_curve_optimal}
\end{figure}

\paragraph{Illustration}
We apply this method to compare the RL-learned policy from the previous section to the clinician policy.  Figure~\ref{fig:u_curve_optimal} plots the average mortality rate of patients as a function of the difference between the dosage recommended by a learned optimal policy and the actual dosage given to the patients. Both plots demonstrate a characteristic `U' shape, with the lowest mortality attained for zero dosage differences, indicating that patients are more likely to survive the closer the treatment they are given is to the treatment the learned policy would have recommended. This observation appears to support the utility of the learned policy. 

\begin{figure}[ht]
	\centering
	\begin{subfigure}{0.45\linewidth}
		\centering
		\includegraphics[width=\linewidth]{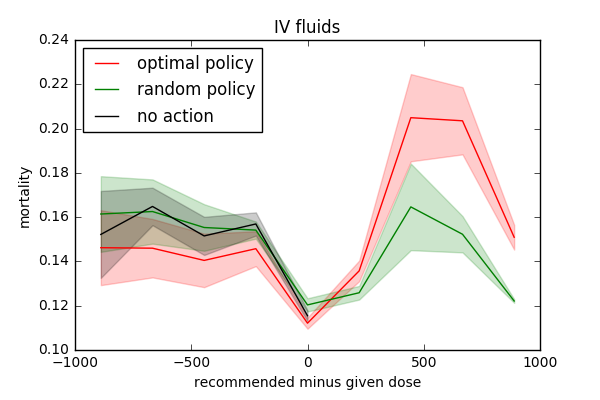}
	\end{subfigure}
	\rulesep
	\begin{subfigure}{0.45\linewidth}
		\centering
		\includegraphics[width=\linewidth]{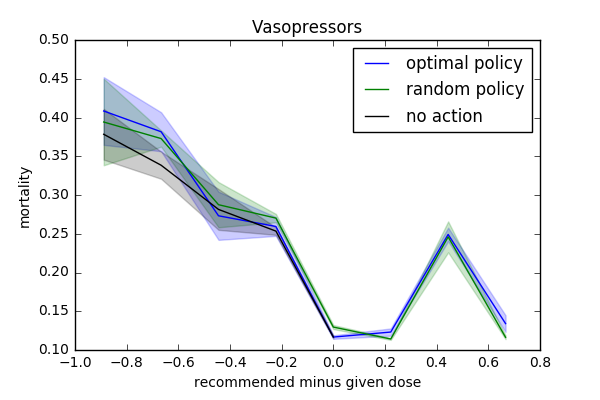}
	\end{subfigure}
	\caption{Mortality vs. deviation from the dose recommended by the optimal policy learned from the model, as well as a random policy and a no-action policy. The performances of all policies look remarkably similar. The binning of actions into dosage quartiles (see also Figure~\ref{fig:vasopressors_hist}) results in the learned policy's dosage recommendation be insignificant for most data-points in these plots. }
	\label{fig:u_curve_all_policies}
\end{figure}

However, upon a closer inspection of the data we see that such an observation can easily be an artifact of confounding factors and the way actions are binned.  In Figure~\ref{fig:u_curve_all_policies} we plot the same data as in Figure~\ref{fig:u_curve_optimal}, but overlay it with similar plots for two obviously suboptimal treatment policies: a policy which assigns a completely random action to every patient, and a policy in which patients never receive any treatment. The plots exhibit the same U-shaped curves even for two obviously bad policies, and in the vasopressors case the results for all three policies are indistinguishable.

\begin{figure}[ht]
	\centering
	\begin{subfigure}{0.45\linewidth}
		\centering
		\includegraphics[width=\linewidth]{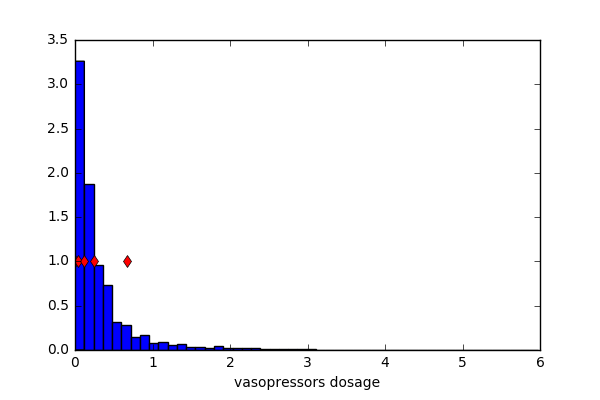}
	\end{subfigure}
	\caption{Histogram of dosages recommended by clinicians'. The recommended dosages have a long tail, making the medians of all but the last quartiles very close on the scale of possible dosages.}
	\label{fig:vasopressors_hist}
\end{figure}

Why does this happen?  Let us consider the no-treatment policy.  The u-curve for this policy simply demonstrates the correlation between dosage given to patients and mortality. This correlation is intuitive: sicker patients with a higher risk of mortality are given higher treatment dosages.  The remaining u-curves appear to be similar to the no-action policy because of the way continuous actions (dosages) are binned into discrete values. For every treatment, one bin represents no treatment, and the other four bins represent four quartiles of the dosages in the data. Because the given dosage is exponentially distributed (see Figure~\ref{fig:vasopressors_hist}), the medians dosages (red diamonds in Figure~\ref{fig:vasopressors_hist}) for all but the last quartile are negligible on the scale of the x-axes of the u-curves, and therefore for this binning of the actions, the u-curves are replicas of the no-action curves with the exception of patients for whom the maximal dose possible was recommended.

\paragraph{Diagnosing and Mitigating Concerns}
It is possible to try and avoid this affect by choosing bins of equal sizes rather than quantiles and focusing only on a narrower range of actions, but care must be taken since in that case nearly all of the observations will be assigned to the first few bins, and the observations for the more severe patients will be very sparse.  In general, we recommend that it is best to avoid these ad-hoc methods; as with the question of learning a state respresentation, when quantiative evaluation is challenging we should turn to feedback and checks of face-validity from expert clinicians and the clinical literature.

\section{Recommendations for Researchers}

Given the challenges above, what are we to do?  Retrospective critical care data sets such as the one we described are a gold mine of information, and to dismiss them entirely would be the equivalent of telling a clinician to avoid learning from their colleagues and focus only on their own experience.  Just as it is unethical to present sloppy research results, it is also unethical to not leverage data that could improve patient health.  Below we summarize the issues discussed in this paper and provide several recommendations for researchers working on questions involving optimizing sequential decisions in the healthcare domain.

\paragraph{Design data collection and representation to support causal conclusions.} To justify clinical adoption of policies learned from observational data, it is critical that any causal conclusions are supported by the data used by the learning algorithm. This is a process that should be guided by domain knowledge, as claims that all confounding variables have been accounted for cannot be made based on statistical quantities alone. Instead, researchers should strive to collaborate with practitioners to verify that any information that guides current practice has been sufficiently well recorded. In addition, it is equally important that any model or variable selection that follows data collection preserves such information. Efforts to reduce variance through dimensionality reduction may introduce confounding bias if confounders are left out of the analysis.

\paragraph{Limit yourself to actions that were taken and policies that are similar to physicians'.} At its core, we cannot evaluate things that we have not tried.  It is easy for algorithms to become overly optimistic (or in some cases, pessimistic) about actions rarely performed in practice on particular patients.  Therefore, all studies should limit themselves to decisions that have sufficient data to be evaluated.  Conceptually, this is quite reasonable.  We cannot expect, from retrospective data, for an AI to identify an amazing treatment policy that was never tried by clinicians but somehow ``hiding'' in the data: AI is not magic. However, we can use AI to identify which decisions, of those taken, were best, and thus create a decision support tool that helps clinicians mimic the best doctors at their best moments. To take that practice of caution a step further, we should try to evaluate policies which are not too dissimilar from physicians, and could be evaluated more accurately using statistical tools such as importance sampling estimators.

\paragraph{Combine an understanding of the limitations of AI algorithms and evaluation methods to calibrate expectations and define goals accordingly.}
More broadly, if we understand the limitations of existing methods, we can use AI algorithms to perform more modest tasks which could still be of tremendous use in medicine. For example, current algorithms using available data are well suited to build treatment recommendation systems or suggest small changes to existing policies which are worth exploring further.

\paragraph{Be cognizant of effective sample sizes.} Most state of the art off-policy evaluation methods today use some sort of importance sampling estimate. As discussed in section \ref{sec:IS_estimators}, when we use these methods to evaluate deterministic policies, the number of sequences with non-zero weights decays exponentially with the length of sequences. As a result, it is very common for the effective number of samples used to evaluate a policy to be only a small fraction of the actual cohort size. We should therefore make sure the effective sample sizes used to evaluate policies is large enough for our evaluations to be statistically significant. Limiting ourselves to policies that are similar to physicians', as discussed above, will also be beneficial for increasing the effective sample size.

\paragraph{Interpret the features that are being used by the policy, and also where the policies differ from what was done.}
Finally, as we have alluded many times, not all problems can be solved by fancier RL algorithms or fancier RL evaluation techniques.  Currently, RL algorithms usually operate as black boxes---they receive a set of data as input, and output a policy. These policies are often difficult to interpret and it is usually hard to discern which features in the data resulted in the specific suggestion for an action. Uninterpretable policies are of concern not only because they do not allow experts the opportunity to identify errors and artifacts, but also because they can slow adoption as clinicians are (rightly) unlikely to attempt treatments lacking in clinical reasoning.

Recent works have recently addressed the issue of modifying ML algorithms to produce more interpretable predictions~\citep{vellido2012making,doshi2017towards,lipton2016mythos,wang2015falling} and some have tackled the specific problem of interpretability of ML in the healthcare domain~\citep{caruana2015intelligible}. More recently some works have also specifically attempted to address the problem of interpretability in RL algorithms~\citep{maes2012policy,hein2017particle}. There are a variety of ways to interrogate policies suggested by agents: we can consider what are the input features that most affect the output recommendation, and do changing the inputs change the recommendation in expected ways?  What features are most predictive of there being a difference between the recommendation and the clinician's action? Dependence on inputs that are clearly irrelevant from a clinical perspective can be a useful signal that something is wrong.

Finally, expert evaluation should be performed not only on the final policy but at every stage of the learning process. For example, the representation of states can be evaluated to check that states represent meaningful sub-groups in the population. Domain expert input is important in setting up the RL problem, in particular in choosing the reward scheme. The performance of all RL algorithms will depend on what outcome they are designed to optimize for, and thus clinicians should weigh in on questions such as whether short term goals such as keeping patients' vitals within a certain range will translate into desirable long-term outcomes such as mortality.  RL is a powerful tool which can allow us to improve and refine existing treatment policies, but clinical experts must still be called upon to exercise their judgment and expertise.

\section{Conclusions}
This article raises more questions than it answers: evaluation, especially of sequential decisions, on retrospective data is a fundamentally hard problem.  We hope that by exposing these issues, we create an informed community of researchers who understand both the promise and the challenges of working with such data.  We hope that readers will be inspired to think deeply about careful evaluation, and we spark a conversation on best practices for research in this area.

\bibliographystyle{plainnat}
\bibliography{RL_OPE_review}

\begin{thebibliography}{20}
\providecommand{\natexlab}[1]{#1}
\providecommand{\url}[1]{\texttt{#1}}
\expandafter\ifx\csname urlstyle\endcsname\relax
  \providecommand{\doi}[1]{doi: #1}\else
  \providecommand{\doi}{doi: \begingroup \urlstyle{rm}\Url}\fi

\bibitem[Caruana et~al.(2015)Caruana, Lou, Gehrke, Koch, Sturm, and
  Elhadad]{caruana2015intelligible}
Rich Caruana, Yin Lou, Johannes Gehrke, Paul Koch, Marc Sturm, and Noemie
  Elhadad.
\newblock Intelligible models for healthcare: Predicting pneumonia risk and
  hospital 30-day readmission.
\newblock In \emph{Proceedings of the 21th ACM SIGKDD International Conference
  on Knowledge Discovery and Data Mining}, pages 1721--1730. ACM, 2015.

\bibitem[Chakraborty and Murphy(2014)]{chakraborty2014dynamic}
Bibhas Chakraborty and Susan~A Murphy.
\newblock Dynamic treatment regimes.
\newblock \emph{Annual review of statistics and its application}, 1:\penalty0
  447--464, 2014.

\bibitem[Doshi-Velez and Kim(2017)]{doshi2017towards}
Finale Doshi-Velez and Been Kim.
\newblock Towards a rigorous science of interpretable machine learning.
\newblock 2017.

\bibitem[Hein et~al.(2017)Hein, Hentschel, Runkler, and
  Udluft]{hein2017particle}
Daniel Hein, Alexander Hentschel, Thomas Runkler, and Steffen Udluft.
\newblock Particle swarm optimization for generating interpretable fuzzy
  reinforcement learning policies.
\newblock \emph{Engineering Applications of Artificial Intelligence},
  65:\penalty0 87--98, 2017.

\bibitem[Imbens and Rubin(2015)]{imbens2015causal}
Guido~W Imbens and Donald~B Rubin.
\newblock \emph{Causal inference in statistics, social, and biomedical
  sciences}.
\newblock Cambridge University Press, 2015.

\bibitem[Jiang and Li(2015)]{jiang2015doubly}
Nan Jiang and Lihong Li.
\newblock Doubly robust off-policy value evaluation for reinforcement learning.
\newblock \emph{arXiv preprint arXiv:1511.03722}, 2015.

\bibitem[Johnson et~al.(2017)Johnson, Stone, Celi, and
  Pollard]{johnson2017mimic}
Alistair~EW Johnson, David~J Stone, Leo~A Celi, and Tom~J Pollard.
\newblock The mimic code repository: enabling reproducibility in critical care
  research.
\newblock \emph{Journal of the American Medical Informatics Association}, page
  ocx084, 2017.

\bibitem[Lipton(2016)]{lipton2016mythos}
Zachary~C Lipton.
\newblock The mythos of model interpretability.
\newblock \emph{arXiv preprint arXiv:1606.03490}, 2016.

\bibitem[Maes et~al.(2012)Maes, Fonteneau, Wehenkel, and Ernst]{maes2012policy}
Francis Maes, Raphael Fonteneau, Louis Wehenkel, and Damien Ernst.
\newblock Policy search in a space of simple closed-form formulas: towards
  interpretability of reinforcement learning.
\newblock In \emph{International Conference on Discovery Science}, pages
  37--51. Springer, 2012.

\bibitem[Pearl(2009)]{pearl2009causality}
Judea Pearl.
\newblock \emph{Causality}.
\newblock Cambridge university press, 2009.

\bibitem[Prasad et~al.(2017)Prasad, Cheng, Chivers, Draugelis, and
  Engelhardt]{prasad2017reinforcement}
Niranjani Prasad, Li-Fang Cheng, Corey Chivers, Michael Draugelis, and
  Barbara~E Engelhardt.
\newblock A reinforcement learning approach to weaning of mechanical
  ventilation in intensive care units.
\newblock \emph{arXiv preprint arXiv:1704.06300}, 2017.

\bibitem[Precup et~al.(2000)Precup, Sutton, and Singh]{precup2000eligibility}
Doina Precup, Richard~S Sutton, and Satinder~P Singh.
\newblock Eligibility traces for off-policy policy evaluation.
\newblock In \emph{ICML}, pages 759--766. Citeseer, 2000.

\bibitem[Raghu et~al.(2017)Raghu, Komorowski, Ahmed, Celi, Szolovits, and
  Ghassemi]{raghu2017deep}
Aniruddh Raghu, Matthieu Komorowski, Imran Ahmed, Leo Celi, Peter Szolovits,
  and Marzyeh Ghassemi.
\newblock Deep reinforcement learning for sepsis treatment.
\newblock \emph{arXiv preprint arXiv:1711.09602}, 2017.

\bibitem[Robins(2000)]{robins2000robust}
James~M Robins.
\newblock Robust estimation in sequentially ignorable missing data and causal
  inference models.
\newblock In \emph{Proceedings of the American Statistical Association}, volume
  1999, pages 6--10, 2000.

\bibitem[Shalit et~al.(2016)Shalit, Johansson, and
  Sontag]{shalit2016estimating}
Uri Shalit, Fredrik Johansson, and David Sontag.
\newblock Estimating individual treatment effect: generalization bounds and
  algorithms.
\newblock \emph{arXiv preprint arXiv:1606.03976}, 2016.

\bibitem[Shortreed et~al.(2011)Shortreed, Laber, Lizotte, Stroup, Pineau, and
  Murphy]{shortreed2011informing}
Susan~M Shortreed, Eric Laber, Daniel~J Lizotte, T~Scott Stroup, Joelle Pineau,
  and Susan~A Murphy.
\newblock Informing sequential clinical decision-making through reinforcement
  learning: an empirical study.
\newblock \emph{Machine learning}, 84\penalty0 (1-2):\penalty0 109--136, 2011.

\bibitem[Singer et~al.(2016)Singer, Deutschman, Seymour, Shankar-Hari, Annane,
  Bauer, Bellomo, Bernard, Chiche, Coopersmith, et~al.]{singer2016third}
Mervyn Singer, Clifford~S Deutschman, Christopher~Warren Seymour, Manu
  Shankar-Hari, Djillali Annane, Michael Bauer, Rinaldo Bellomo, Gordon~R
  Bernard, Jean-Daniel Chiche, Craig~M Coopersmith, et~al.
\newblock The third international consensus definitions for sepsis and septic
  shock (sepsis-3).
\newblock \emph{Jama}, 315\penalty0 (8):\penalty0 801--810, 2016.

\bibitem[Thomas and Brunskill(2016)]{thomas2016data}
Philip Thomas and Emma Brunskill.
\newblock Data-efficient off-policy policy evaluation for reinforcement
  learning.
\newblock In \emph{International Conference on Machine Learning}, pages
  2139--2148, 2016.

\bibitem[Vellido et~al.(2012)Vellido, Mart{\'\i}n-Guerrero, and
  Lisboa]{vellido2012making}
Alfredo Vellido, Jos{\'e}~David Mart{\'\i}n-Guerrero, and Paulo~JG Lisboa.
\newblock Making machine learning models interpretable.
\newblock In \emph{ESANN}, volume~12, pages 163--172. Citeseer, 2012.

\bibitem[Wang and Rudin(2015)]{wang2015falling}
Fulton Wang and Cynthia Rudin.
\newblock Falling rule lists.
\newblock In \emph{Artificial Intelligence and Statistics}, pages 1013--1022,
  2015.

\end{thebibliography}

\end{document}